%% file: main.tex
\documentclass[10pt,twocolumn,letterpaper]{article}

\usepackage{iccv}
\usepackage{times}
\usepackage{epsfig}
\usepackage{graphicx}
\usepackage{amsmath}
\usepackage{amssymb}

\usepackage{kotex}
\usepackage{braket}
\usepackage{color}
\usepackage{xcolor}
\usepackage{cases}
\usepackage{nicefrac}
\usepackage{tabu}
\usepackage{booktabs}
\usepackage{multirow}
\usepackage{subcaption}
\usepackage{caption}
\usepackage{soul}
\usepackage[normalem]{ulem}
\usepackage{stfloats}

\usepackage[breaklinks=true,bookmarks=false]{hyperref}

\iccvfinalcopy 

\def\iccvPaperID{2904} 

\begin{document}

\title{Cross-modal Variational Auto-encoder with \\Distributed Latent Spaces and Associators}

\author{
Dae Ung Jo\textsuperscript{1}
\,
ByeongJu Lee\textsuperscript{1}
\,
Jongwon Choi\textsuperscript{2} 
\,
Haanju Yoo\textsuperscript{3}
\,
Jin Young Choi\textsuperscript{1}
\\
{\tt\small \{mardaewoon,adolys,jychoi\}@snu.ac.kr, \{jw17.choi,haanju.yoo\}@samsung.com} 
\\
\textsuperscript{1}Department of ECE, ASRI, Seoul National University, Korea\\
\textsuperscript{2}Samsung SDS, Korea \, \textsuperscript{3}Samsung Research, Korea\\
}

\maketitle
\begin{abstract}
\input{0-Abstract}
\end{abstract}

\section{Introduction}
\input{1-Introduction_new.tex}

\section{Related Works}
\input{2-Related}

\section{Method}
\input{3-1-Method_intro}

\input{3-2-Method_bayes}

\input{3-3-Method_network}

\input{3-4-Method_compare}

\section{Experiment}
\input{4-Experiment}

\section{Conclusion}
\input{5-Conclusion}

{\small
\bibliographystyle{ieee}
\bibliography{refs}
}

\end{document}

%% file: 0-Abstract.tex
In this paper, we propose a novel structure for a cross-modal data association, which is inspired by the recent research on the associative learning structure of the brain.
We formulate the cross-modal association in Bayesian inference framework realized by a deep neural network with multiple variational auto-encoders and variational associators.
The variational associators transfer the latent spaces between auto-encoders that represent different modalities. 
The proposed structure successfully associates even heterogeneous modal data and easily incorporates the additional modality to the entire network via the proposed cross-modal associator.
Furthermore, the proposed structure can be trained with only a small amount of paired data
since auto-encoders can be trained by unsupervised manner.
Through experiments, the effectiveness of the proposed structure is validated on various datasets including visual and auditory data.

%% file: 1-Introduction_new.tex
\input{figure_teasor.tex}
The brain combines multisensory information to understand the surrounding situation.
Through various sensory experiences, humans learn the relationships between multisensory data and understand the experienced situation. 
This mechanism of learning the relationship between different stimuli is called associative learning
~\cite{1993_synaptic_memory,1998_cortical_plasticity,1927_Pavlov,1928_Pavlov,1999_neurogenesis_mice,1997_elementary_associative,2004_auditory_cortex}. 
Because of the associative learning mechanism, humans can robustly understand and perceive their surrounding situations even when only some of the modalities are available.

In the field of machine learning, utilizing multi-modality is also important issues because of its usefulness in a wide range of applications~\cite{2018_multimodal_survey, 2013_representation_survey}.
As a representative example, object recognition and scene understanding methods based on multi-modal data outperform the methods using only single-modal data~\cite{2016_Temporal,2011_MultimodalDeep}.
Moreover, one can generate the synthesized data for a missing or desired modality~\cite{2016_Robotscene,2018_lim_pose,2018_Localize_sound,2018_Handpose,2018_MixandMatch,2017_Yoo_Regression}.
The cross-modal data association is one of the fundamental steps to understand the relationships among multi-modal data.

For this reason, a lot of studies have attempted to solve this cross-modal data association problem with a deep learning algorithm~\cite{2018_multimodal_survey}. 
Audio data and visual data has been associated in 
\cite{2016_Temporal,2018_Localize_sound, 2012_multimodal_boltzman, 2011_MultimodalDeep}. The works in \cite{2018_Handpose, 2017_handpose_crossing} deal with visual data and hand pose data. 
The association between heterogeneous visual data has been tried in \cite{2016_Robotscene,2018_MixandMatch}, where the heterogeneous visual data includes RGB image, depth image, and segmentation map.
Recent studies have adopted an approach that encodes cross-modal data into a shared latent space to memorize common features among multiple modalities 
~\cite{2016_Robotscene,2016_Temporal,2011_MultimodalDeep,2018_Handpose, 2018_multimodal_weakly}.

However, as pointed out in \cite{2017_conditional_multimodal_embedding}, most existing studies did not consider the case that the characteristic of each modality is very different from others. In this case, it is hard to encode meaningful common feature representing all characteristics of the heterogeneous modalities or the encoding could be biased to a dominant modality. Furthermore, the existing approach encounters a scalability problem since the capacity of the shared latent space will decrease as the number of modalities increases.

To mitigate the limitation of shared latent space approach, we propose an approach that adopts a distributed latent space concept.
In our approach, as shown in Figure \ref{fig:teaser}, each modality is encoded by the usual variational auto-encoder (VAE) and the distributed latent space encoded from each modality is associated with the other modality via cross-modal associator between them. 
This approach is inspired by the research on associative memory cells in the brain~\cite{2018_associative_brain, 2017_associative_brain}, where the intra-modal association is done in each sensory cortex and the cross-modal association is done among sensory cortices.

The intra-modal association is realized by VAE and the cross-modal association is realized by the proposed {\it associator}.
In the proposed structure, the information on each modality is memorized in each VAE and the cross-modal association can be easily performed through associator between them. The loss function to train the model is derived by the variational inference framework. The advantages of the proposed approach are discussed in view of generalization ability for semi-supervised learning, scalability, and flexibility of encoding dimension. 
In experiments, the effectiveness, performance, and advantages of the proposed approach are evaluated through comparison with the existing methods and empirical self-analysis using various datasets including voice and visual data.

%% file: figure_teasor.tex
\begin{figure}
\centering 
\includegraphics[width=0.85\linewidth]{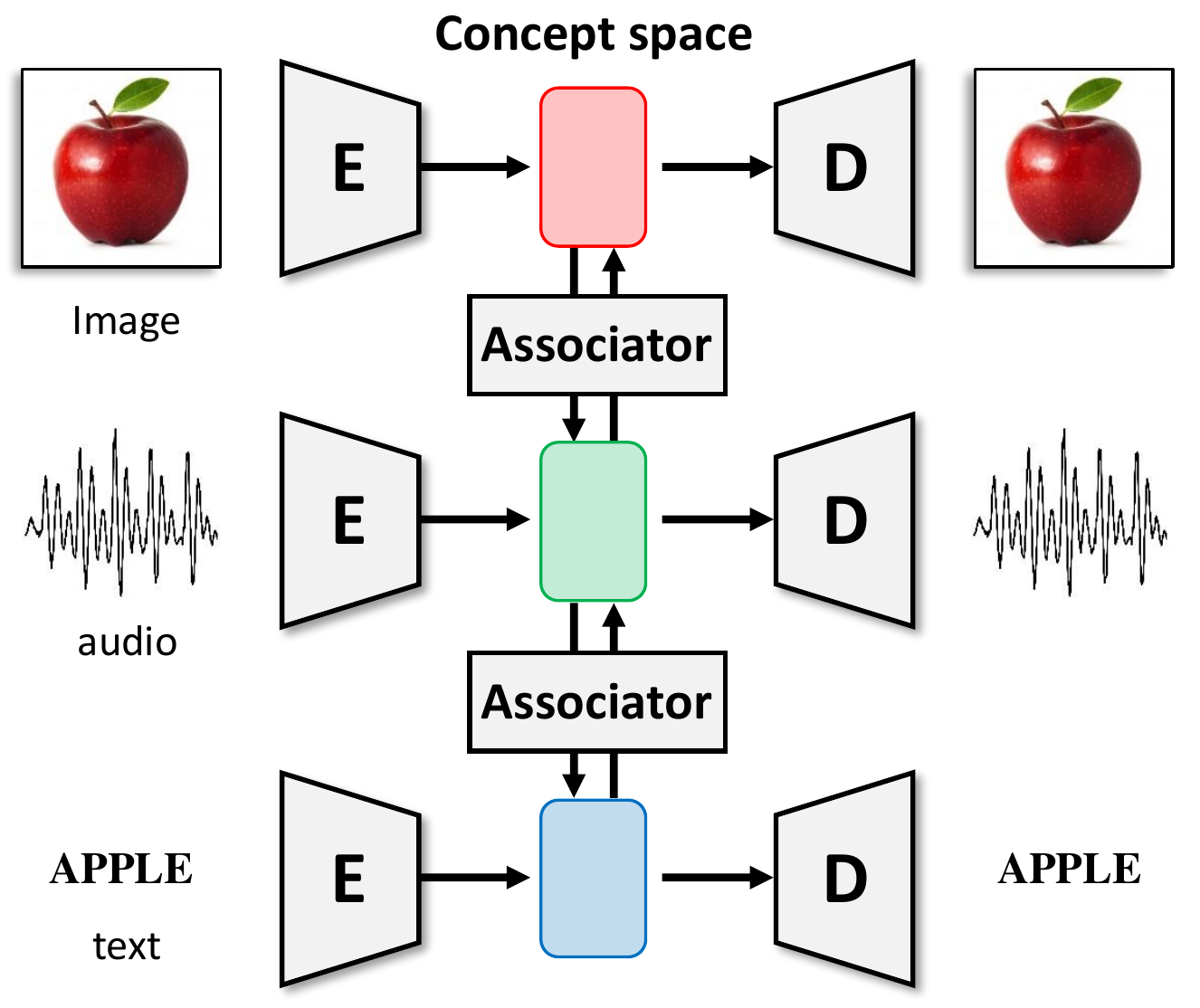}
\caption{
Conceptual illustration of the proposed cross-modal association network.
The proposed network has modality-specific encoders and decoders for each modality (image, audio, text).
Each modality has its own latent space, which is painted with a different color (red, green, blue).
The latent spaces are connected via the proposed \textit{associator} which associates two different modalities.
}
\label{fig:teaser}
\end{figure}

%% file: 2-Related.tex
\subsection{Multi-modality in Machine Learning}

One of the major issues in machine learning is exploiting multi-modal data for various applications, such as data generation~\cite{2013_VAE,2014_GAN, 2018_Handpose, 2017_Yoo_Regression}, retrieval~\cite{2016_retrieval_survey} and recognition~\cite{2016_Temporal,2011_MultimodalDeep}. 
There are a lot of studies that extract modality independent features by finding the shared representation of multi-modal data~\cite{2018_multimodal_survey}.
The shared representation is utilized in diverse applications such as handling a missing modality~\cite{2016_Robotscene,2018_Handpose} or accomplishing better performance than models trained on single-modal data~\cite{2016_Temporal,2011_MultimodalDeep}.
The research related to multi-modality can be categorized into two groups~\cite{2018_multimodal_survey}.

One is a method mapping data from diverse modalities to the shared latent space.
\cite{2018_multimodal_weakly} proposes the extended version of a Variational auto-encoder~\cite{2013_VAE} which combines distribution parameters from encoders and calculate integrated distribution parameters.
\cite{2018_Handpose} also a variant of Variational auto-encoder~\cite{2013_VAE} for hand pose estimation with multi-modal data. The model proposed by \cite{2018_Handpose} chooses the input modality and the output modality pair and train the corresponding encoder and decoder pair at every iteration.
\cite{2016_Robotscene} trains an auto-encoder that takes RGB images, depth images and semantic images as its network input, then the trained model can a generate complete depth image and semantic image from an RGB image and partial depth and semantic image.
\cite{2011_MultimodalDeep} builds a deep-belief network structure that maps audio data and lip images into the common hidden node for audio-visual speech recognition. 
\cite{2016_Temporal} extends the RBM structure to reflect the sequential characteristic of a speech dataset.

The other group comprises methods that encode the corresponding data to the latent space of each modality but enforce similarity constraints to corresponded latent vectors.
\cite{2018_MixandMatch} trains domain specific encoders and decoders, allowing encoders and decoders from different modality to be combined, then, the model is able to generate an unseen data pair by combining the encoders and decoders.
\cite{2017_conditional_multimodal_embedding} extracts low-level representation from original data first. Then \cite{2017_conditional_multimodal_embedding} trains auto-encoders for each modality and enforces similarity constraints to embedding spaces of each auto-encoders for correlated data pair.
In \cite{2013_DeViSE}, a model is trained to maximize the similarity of an image feature and a vectorized label to infer a proper label for a given image.

\subsection{Associative Learning inspired by the Brain}

The artificial neural network is an engineering model inspired by the biological mechanism of the brain.
Parameters of those networks are usually updated by Hebbian learning rule where weight connections between firing nodes for input data are strengthened~\cite{1961_Hebbian}.
The Hopfield network and Boltzmann machine are representative examples~\cite{1985_Boltzmann_Machine}. 
The Hopfield network models associative memory of human, thus network is trained to memorize specific patterns.
Even if the input is incomplete, The Hopfield network can restore incomplete data through recurrent iteration.
The Boltzmann machine is a stochastic version of the Hopfield network, which can learn a latent representation for input data through its hidden nodes.

There have been many studies that investigate associative learning from the perspective of neuroscience~\cite{1993_synaptic_memory, 1998_cortical_plasticity, 1999_neurogenesis_mice, 2004_auditory_cortex}.
In the recent study which tried to analyze associative learning at the cellular substrate level~\cite{2018_associative_brain, 2017_associative_brain, 1997_elementary_associative}, they introduce the associative memory cells to describe brain neurons which are mainly involved in integration and storage of associated signals.
A brain learns associated information by enhancing the strength of the synapses between co-activated associative memory cells activated by associated signals.
In this paper, we realize the cross-modal association mechanism recently proposed in \cite{2018_associative_brain}, which assumes the comprehensive diagram based on the associative memory cells.

%% file: 3-1-Method_intro.tex
\subsection{Problem Statements}
\label{sec:problem_statement}

According to recent studies~\cite{2017_associative_brain, 2018_associative_brain}, associative learning process in the brain includes intra-modal and cross-modal association processes.
The intra-modal association process is to make humans familiar with single-modal sensory information.
On the other hand, the cross-modal association process is accomplished to enhance the strength of the synapses connecting multi-modal information to be associated.
The goal of this paper is to establish the Bayesian formulation of these two association processes and to realize them in a variational auto-encoder framework.

%% file: 3-2-Method_bayes.tex
\subsection{Graphical Model of Cross-Modal Association}

\input{figure_graphical_model.tex}

\subsubsection{Intra-Modal Association}

Intra-modal association is the process of memorizing single-domain information.
To efficiently memorize a vast amount of information, the model needs to extract the expressive features of the data.
One way to make the encoding model remember the features of the data in an unsupervised manner, is to formulate a mathematical model reconstructing the original sensory data from the encoded information.
Figure~\ref{fig:graphical structure}.(a) shows Bayesian graphical model to formulate the intra-modal association to memorize a distribution of the latent variable $\mathbf{z}_i$ associated with the input variable $\mathbf{x}_i$ for an observation in modality $i$.
In the Bayesian framework, the objective is to infer model parameter $\theta_i$ of posterior distribution $p_{\theta_i}(\mathbf{z}_i|\mathbf{x}_i)$.

One of the most popular approaches to approximate an intractable posterior is the variational inference method.
In this method, the variational distribution $q_{\phi_i}(\mathbf{z}_i|\mathbf{x}_i)$ approximates the true posterior $p_{\theta_i}(\mathbf{z}_i|\mathbf{x}_i)$ by minimizing the Kullback-Leibler divergence, $D_{KL} \left(q_{\phi_i}(\mathbf{z}_i|\mathbf{x}_i) \| p_{\theta_i}(\mathbf{z}_i|\mathbf{x}_i)\right)$.
According to \cite{2013_VAE}, the minimization of $D_{KL} \left(q_{\phi_i}(\mathbf{z}_i|\mathbf{x}_i) \| p_{\theta_i}(\mathbf{z}_i|\mathbf{x}_i)\right)$ can be replaced with the maximization of the evidence lower bound, given by
\begin{equation}
\begin{aligned}
\label{eq:vae_elbo}
    \mathcal{L}(q_{\phi_i}(\mathbf{z}_{i}|\mathbf{x}_i)) = &-
    D_{KL}(q_{\phi_i}(\mathbf{z}_i|\mathbf{x}_i) \| p_{\theta_i}(\mathbf{z}_i))\\
    &+ \mathbb{E}_{q_{\phi_i}(\mathbf{z}_i|\mathbf{x}_i)}[\log{p_{\theta_i}(\mathbf{x}_i|\mathbf{z}_i)}], \\
\end{aligned}
\end{equation}
where $\mathbb{E}_{q_{\phi_i}(\mathbf{z}_i|\mathbf{x}_i)}$ indicates expectation over distribution $q_{\phi_i}(\mathbf{z}_i|\mathbf{x}_i)$.

\input{figure_network.tex}

\subsubsection{Cross-Modal Association}

In this section,
we design a graphical model to represent the cross-modal association mechanism as in Figure~\ref{fig:graphical structure}.(b) Without loss of generality, we consider a path from modality $i$ to $j$.
From observations of an associated variable pair $(\mathbf{x}_i, \mathbf{x}_j)$, the distribution parameter $\rho_{ji}$ is inferred to model the association between $\mathbf{z}_i$ and $\mathbf{z}_j$.

For a given observation pair $(\mathbf{x}_i, \mathbf{x}_j)$,
the cross-posterior distribution $p_{\theta_{i}, \rho_{ji}}(\mathbf{z}_j|\mathbf{x}_i)$ is defined by marginalization for $\mathbf{z}_i$ as 
\begin{equation}
\begin{aligned}
	p_{\theta_{i}, \rho_{ji}}(\mathbf{z}_{j}|\mathbf{x}_i) = \int p_{\rho_{ji}}(\mathbf{z}_j|\mathbf{z}_i) p_{\theta_{i}}(\mathbf{z}_i|\mathbf{x}_i) \: d\mathbf{z}_i.
\end{aligned}
\label{eq:cross_posterior}
\end{equation}
To establish the cross-modal association model, we define a variational distribution for cross-posterior distribution
$q_{\phi_i, \rho_{ji}}(\mathbf{z}_{j}|\mathbf{x}_i)$.
Then, to infer the distribution parameters $(\phi_i, \rho_{ji})$, we minimize Kullback-Leibler divergence between $p_{\theta_j}(\mathbf{z}_j|\mathbf{x}_j)$ and $q_{\phi_i, \rho_{ji}}(\mathbf{z}_{j}|\mathbf{x}_i)$. To avoid clutter, subscripts for the distribution parameters are omitted in the remainders of this section.
Kullback-Leibler divergence between $p(\mathbf{z}_j|\mathbf{x}_j)$ and $q(\mathbf{z}_{j}|\mathbf{x}_i)$ is given by
\begin{equation}
	D_{KL}(q(\mathbf{z}_j|\mathbf{x}_i) \| p(\mathbf{z}_j|\mathbf{x}_j)) 
    = \log{p(\mathbf{x}_j)} - \mathcal{L}(q(\mathbf{z}_{j}|\mathbf{x}_i)), 
\end{equation}
where
\begin{equation}
    \mathcal{L}(q(\mathbf{z}_{j}|\mathbf{x}_i)) = \int q(\mathbf{z}_{j}|\mathbf{x}_i) \log{\frac{p(\mathbf{x}_j)p(\mathbf{z}_j|\mathbf{x}_j)}{q(\mathbf{z}_{j}|\mathbf{x}_i)}} \: d\mathbf{z}_j.
\label{eq:cross1}
\end{equation}
Since log-evidence $\log{p(\mathbf{x}_j)}$ is independent to the model parameter, the target problem is identical to maximizing the evidence lower bound $\mathcal{L}(q(\mathbf{z}_{j}|\mathbf{x}_i))$.
With probabilistic tricks, $\mathcal{L}(q(\mathbf{z}_{j}|\mathbf{x}_i))$ can be decomposed as following.
\begin{equation}
\begin{aligned}
	\mathcal{L}(q(\mathbf{z}_{j}|\mathbf{x}_i)) =
    &-D_{KL}(q(\mathbf{z}_{j}|\mathbf{x}_i) \| p(\mathbf{z}_j))\\
    &+ \mathbb{E}_{q(\mathbf{z}_{j}|\mathbf{x}_i)}[\log{p(\mathbf{x}_j|\mathbf{z}_j)}].
\end{aligned}
\label{eq:cross2}
\end{equation}
The detail derivation is given in Appendix A of supplementary document.
In Eq.~\eqref{eq:cross2}, the first term is a negative KL divergence term that leads $\mathbf{z}_j$ given by $\mathbf{x}_j$ to have similar distribution with a prior distribution of target modality.
The expectation term in Eq.~\eqref{eq:cross2} minimizes the reconstruction error of decoded output from $\mathbf{z}_j$ fired from $\mathbf{x}_i$, which also promotes the inference for $\rho_{ji}$.
By the similar steps, we can easily derive the opposite association from modality $j$ to modality $i$.

%% file: figure_graphical_model.tex
\begin{figure}
\centering 
\includegraphics[width=0.8\linewidth]{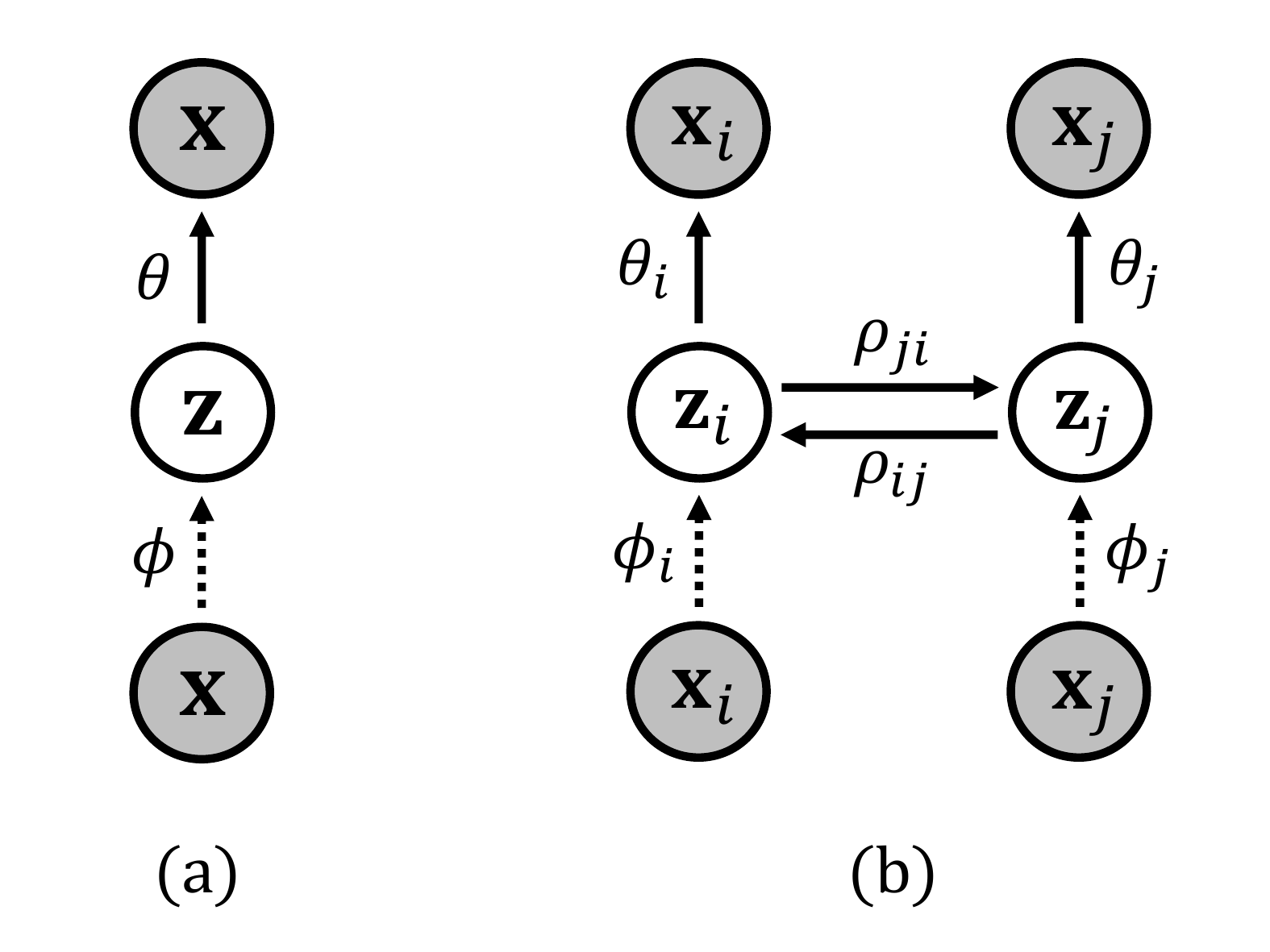}
\caption{
\textbf{Graphical models for intra-modal and cross-modal association.} 
Observable variables are illustrated as shadowed circles.
$\theta, \phi, \rho$ are distribution parameters: $\theta$ for true distribution, $\phi$ for variational distribution, and $\rho$ for cross-modal association model.
Subscripts denote modality.
Dotted lines indicate variational approximation of true probability distribution.
\textbf{(a) Intra-modal association} 
Latent variable $\mathbf{z}$ is obtained by $\mathbf{x}$ through $q_\phi(\mathbf{z}|\mathbf{x})$ and $\mathbf{x}$ is inferred from $\mathbf{z}$ through $p_\theta(\mathbf{x}|\mathbf{z})$
\textbf{(b) Cross-modal association between two modalities} The cross-modal association model has mutual connections between latent variables $\mathbf{z}_i $ and $\mathbf{z}_j$.}
\label{fig:graphical structure}
\end{figure}

%% file: figure_network.tex
\begin{figure*}
\centering 
\includegraphics[width=0.9\linewidth]{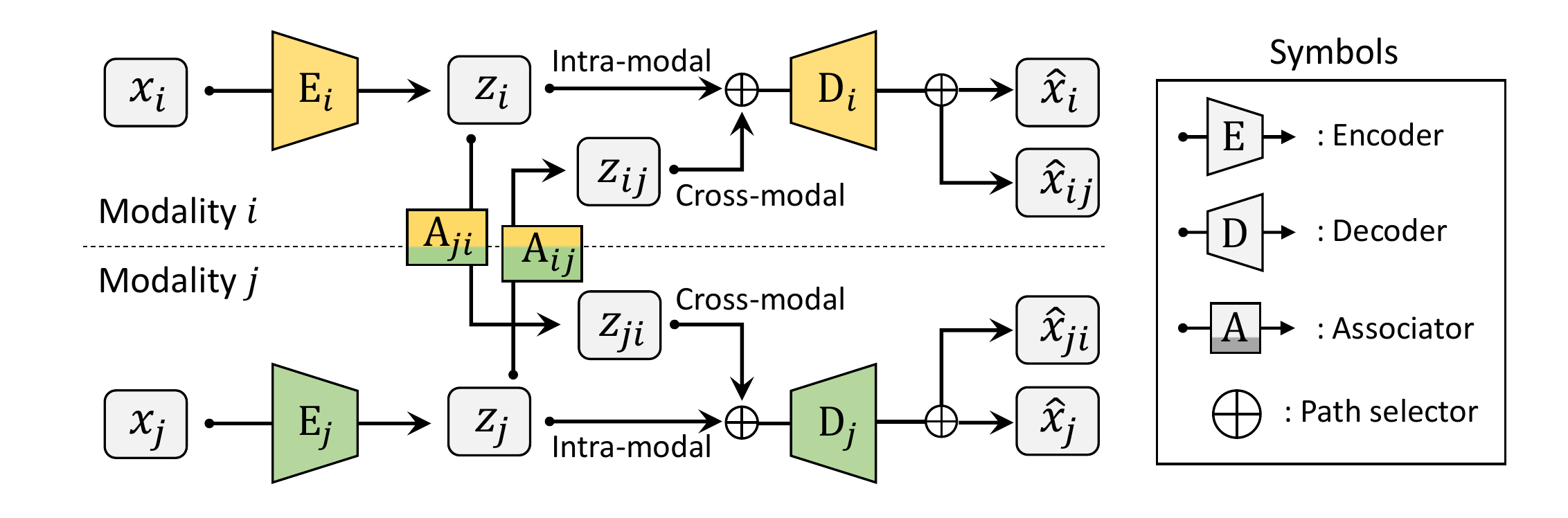}
\caption{Overall structure of the proposed method for the modalities $i$ and  $j$.
For an observation sample $x_i$ for the variable $\mathbf{x}_i$, the intra-modal network of modality $i$ encodes $x_i$ in a latent vector $z_i$ for the latent variable
$\mathbf{z}_i$ through the encoder $E_i$ and decodes $z_i$ to $\hat{x}_i$ through decoder $D_i$.
In the case of association from modality $i$ to $j$, a sample $x_i$ is encoded to $z_{ji}$ through the encoder $E_i$ and the associator $A_{ji}$. Then, $z_{ji}$ is decoded to $\hat{x}_{ji}$ through $D_j$.
The procedure for the opposite direction is performed in the same way.
}
\label{fig:network}
\end{figure*}

%% file: 3-3-Method_network.tex
\subsection{Realization: Cross-Modal Association Network}

We accomplish a realization of the aforementioned intra-modal and cross-modal association models by extending the Variational Auto-Encoder framework (VAE)~\cite{2013_VAE}.
Figure~\ref{fig:network} illustrates the proposed cross-modal association network for modality $i$ and $j$.
Although only two modalities are considered in this paper, the proposed model can be applied to the association among three or more modalities also.
In the proposed structure, the {\it encoder} produces the parameter of $q_{\phi_i}(\mathbf{z}_i|\mathbf{x}_i={x}_i)$, and the {\it  decoder} produces the parameter of $p_{\theta_i}(\mathbf{x}_i|\mathbf{z}_i={z}_i)$. The encoder and decoder are realized by deep neural networks.
Likewise, the latent space associating models $p_{\rho_{ji}}(\mathbf{z}_j|\mathbf{z}_i=z_{i})$ and $p_{\rho_{ij}}(\mathbf{z}_i|\mathbf{z}_j=z_{j})$ are also realized by deep neural networks, which are called by \textit{associator}.
Thus, the intra-modal association network contains several auto-encoders, each of which considers one of the multiple modalities only.
The latent spaces of the auto-encoders are connected by {\it associators} in a pairwise manner, which configure the cross-modal association network.

The proposed network is trained in the two phases: intra-modal training phase and cross-modal training phase. 
In the intra-modal training phase, the auto-encoder in each modality is trained separately by minimizing the approximated version of the negative evidence lower bound in Eq.~\eqref{eq:vae_elbo}~\cite{2013_VAE}.
As derived in \cite{2013_VAE}, variational distributions are assumed by the centered isotropic multivariate Gaussian distribution.
For a given observation sample $x_i$, the encoder $E_i$ produces the mean $\mu_{\phi_i}$ and the variance $\sigma_{\phi_i}$ for a Gaussian distribution of $q_{\phi_i}(\mathbf{z}_i|\mathbf{x}_i=x_i)$. 
Then, the latent vector $z_i$ is
sampled as $z_i=\mu_{\phi_i} + \sigma_{\phi_i}*\epsilon$ and $\epsilon \backsim N(0, I)$.
Similarly, the decoder $D_i$ also produces the mean $\mu_{\theta_i}$ and the variance $\sigma_{\theta_i}$ for a Gaussian distribution of $p_{\theta_i}(\mathbf{x}_i|\mathbf{z}_i=z_i)$. 
Then, the reconstruction vector $\hat{x}_i$ is
sampled as $\hat{x}_i=\mu_{\theta_i} +   \sigma_{\theta_i}*\epsilon$ and $\epsilon \backsim N(0, I)$.

Using the samples, the empirical loss for auto-encoder can be derived as
\begin{equation}
\label{eq:intra_loss}
\begin{aligned}
	\mathcal{L}_{int}&(\theta_i, \phi_i; x_i) = -\mathbb{E}_{q_{\phi_i}(\mathbf{z}_i|x_i)}[\log{p_{\theta_i}(x_i|\mathbf{z}_i)}] \\
    &+ \lambda^{'}_{int} D_{KL}(q_{\phi_i}(\mathbf{z}_i|x_i)||p_{\theta_i}(\mathbf{z}_i)),\\
   &=|| x_i-\hat{x}_i||^2_2\\
    &- \lambda_{int} \sum_k^H (1+\log \sigma_{\phi_i(k)}^2 -\mu_{\phi_i(k)}^2 - \sigma_{\phi_i(k)}^2).
\end{aligned}
\end{equation}
where $\lambda_{int}$ is a user-defined parameter and $H$ is the dimension of the latent variable $\mathbf{z}_i$.
$\mu_{\phi_i(k)}$ and $ \sigma_{\phi_i(k)}^2$ denote the $k$-th element of $\mu_{\phi_i}$ and $ \sigma_{\phi_i}^2$.
The detail derivation presents in Appendix B of supplementary document.

After the convergence of the intra-modal training phase, the following cross-modal training phase proceeds to train the associators while freezing the weights of the auto-encoders.
In the same way as in the intra-modal training phase, for a given observation pair $x_i$ and $x_j$, 
the encoders $E_i$ and $E_j$ produce the latent vectors $z_i$ and $z_j$, respectively. 
In addition, associators $A_{ji}$ and $A_{ij}$ produce the latent vectors $z_{ji}$ and $z_{ij}$ using inputs $z_{i}$ and $z_{j}$, respectively. Thereafter, the decoders $D_i$ and $D_j$ produce
the reconstruction vectors $\hat{x}_{ij}$ and $\hat{x}_{ji}$ from $z_{ij}$ and $z_{ji}$, respectively.

Using the samples, the empirical loss for $A_{ji}$ is designed according to Eq.~\eqref{eq:cross2} as follows:
\begin{equation} 
\label{eq:cross_loss}
\begin{aligned}
	\mathcal{L}_{crs}&(\rho_{ji};x_i, x_j) = 
    -\mathbb{E}_{q_{\rho_{ji}}(\mathbf{z}_{ji}|x_i)}[\log{p_{\theta_{j}}(x_j|\mathbf{z}_{ji})}] \\
    &+ \lambda^{'}_{crs} D_{KL}(q_{\phi_i,\rho_{ji}}(\mathbf{z}_{ji}|x_i)||p_{\theta_{j}}(\mathbf{z}_{ji}))\\
    &=|| x_j-\hat{x}_{ji}||^2_2\\
    &- \lambda_{crs} \sum_k^H (1+\log \sigma_{\rho_{ji}(k)}^2 -\mu_{\rho_{ji}(k)}^2 - \sigma_{\rho_{ji}(k)}^2).
\end{aligned}
\end{equation}
where $\lambda_{crs}$ is a user-defined parameter and $H$ is the dimension of the latent variable $\mathbf{z}_{ji}$.
($\mu_{\rho_{ji}}$, $ \sigma_{\rho_{ji}}^2$) are parameters for Gaussian distribution $q_{\phi_i,\rho_{ji}}(\mathbf{z}_{ji}|x_i)$  produced by $A_{ji}$.
The detail derivation presents in Appendix B of supplementary document.

The loss $\mathcal{L}_{crs}(\rho_{ij};x_i, x_j)$ for $A_{ij}$ is given in the same form except the index. Note that all $\mu$'s and $\sigma$'s in Eq.~\eqref{eq:intra_loss} and Eq.~\eqref{eq:cross_loss} are the functions of weights ($\mathbf{w}$) in encoders, decoders, or associators. Hence, the weights of the proposed network are trained by the negative direction of the gradient of the losses with respect to the weights ($\nabla_{\mathbf{w}}\mathcal{L}(\cdot)$).

%% file: 3-4-Method_compare.tex
\subsection{Advantages of Proposed Method}

Owing to the newly introduced {\it associator}, the proposed model can associate heterogeneous modalities effectively.
Reckless coalescence of heterogeneous data may have a fatal impact on 
associative learning such as the problem that
shared latent vectors can be biased to the dominant modality.
However, in our model, the associator acts as a translator
between heterogeneous modalities and thus the characteristics of each latent space are preserved.
Furthermore, in contrast to the existing models which adopt a shared latent space for the different modalities~\cite{2018_multimodal_weakly, 2018_Handpose,2016_Robotscene, 2011_MultimodalDeep}, our structure can provide a flexible dimensional encoding in each latent space depending on the complexity of each modality.
This provides better cross-modal data association results.

The proposed model easily incorporates additional modalities while maintaining the existing modalities.
That is, a new modality can be added via training of only a new associator between an existing auto-encoder and a new auto-encoder.
Though the associator only associates the new modality with one of the existing modalities, the model can associate the new modality with the rest of the modality by passing through multiple associators.

Finally, in contrast to existing models which always require paired data for cross-modal association, our structure can train the associator with the only small amount of paired data in a semi-supervised manner after learning each auto-encoder using unpaired data independently. Since obtaining paired data for cross-modal association is more expensive than obtaining unpaired data, our model is cost-effective. Furthermore, our model is plausible in that, 
when a person learns a cross-modal association, the paired examples are rarely given by a teacher after the person has become familiar with each modality via self-experience without a teacher.

In the following experiment section, we validate the aforementioned advantages of the proposed structure.

%% file: 4-Experiment.tex
This section presents experimental results validating the effectiveness of the proposed method. 
Through our experiments, we used images and voice recordings sharing common semantic meanings to evaluate our model on the visual and the auditory modalities.
The implementation details for network architectures are provided in Appendix C of the supplementary document.

\subsection{Datasets}
\noindent\textbf{Google Speech Commands (GSC)~\cite{2018_GSC_dataset}}:
As the data for the auditory modality, we used the GSC dataset, which consists of 105,829 audio samples containing utterances of 35 short words.
Each audio sample is one-second-long and encoded with a sampling rate of 16KHz.
Among 35 words, we chose 14 words, including words for each digit ('ZERO' to 'NINE') and four traffic commands ('GO,' 'STOP,' 'LEFT,' 'RIGHT').
The chosen set has 54,239 samples.
We extracted the Mel-Frequency Cepstral Coefficient (MFCC) from each audio clip to generate an audio feature.
MFCC has been widely used in the processing of voice data because it reflects the human auditory perception mechanism well~\cite{2010_MFCC_voicerecognition,2000_MFCC_music,2016_MFCC_heartsound}.
The resulting features are $40\times101$ matrices.
We randomly divided the original dataset into training, validation and test sets at the ratio of 8:1:1.

\vspace{0.3cm}
\noindent\textbf{MNIST~\cite{1998_Lecun_MNIST}}:
We used the MNIST dataset as the corresponding visual data to the GSC for each digit.
The MNIST consists of center-aligned $28 \times 28$ gray-scale images for handwritten digits from 0 to 9. The dataset contains 60k and 10k samples for the training set and testing set, respectively.

\vspace{0.3cm}
\noindent\textbf{Fashion-MNIST (F-MNIST)~\cite{2017_FashionMNIST}}:
We used the F-MNIST dataset as another visual modality which has a similar specification to MNIST dataset.
The F-MNIST consists of center-aligned $28 \times 28$ gray-scale images assigned with a label from 10 kinds of clothing such as T-shirt, Trouser and Sneaker.
The dataset contains 60k and 10k samples for the training set and testing set, respectively.

\vspace{0.3cm}
\noindent\textbf{German Traffic Sign Recognition Benchmark (GTSRB)~\cite{2011_German_traffic}}:
For the visual data that correspond to the traffic commands in GSC, we used the GTSRB dataset, which consists of 51,839 RGB color images illustrating 42 kinds of traffic signals.
In particular, to evaluate the performance on pairs of traffic sign images and voice commands in GSC, we chose four pair sets, where each pair set has  similar semantic meaning, i.e., ('Ahead only,' 'GO'), ('No entry for vehicle,' 'STOP'), ('Turn left and ahead,' 'LEFT'), and ('Turn right and ahead,' 'RIGHT').
Then, to prevent the four signs from occupying the entire latent space, we chose additional sign images such as 'No overtaking,' 'Entry to 30kph zone,' 'Prohibit overweighted vehicle,' 'No-waiting zone,' and 'Roundabout'.
The chosen set includes 10,709 samples.
All of the chosen signs have a circular backboard.
The size of each image varies from $15\times15$ to $250\times250$ pixels for each RGB channel in the original dataset.
In our experiments, we resized all images into $52\times52$.

\input{tables/cross_results}

\input{tables/table_recog_vae.tex}
\subsection{Evaluation Metric}

Since datasets in Section.4.1 have no direct matching relationships, we cannot measure cross-likelihood $p(\mathbf{x}_1|\mathbf{x}_2)$ for paired sample $(\mathbf{x}_1, \mathbf{x}_2)$ used in \cite{2018_multimodal_weakly, 2016_joint_multimodal}.
In our work, we used the reconstruction accuracy as the evaluation metric of the association models. The quality of an image reconstructed by an association model can be a valid measure to evaluate the association model since the quality of the reconstructed image is acceptable to both the human and recognition model.
The reconstruction accuracy was measured by our own recognition networks trained with the original data used in our experiments. 
Table~\ref{tabular:recog_vae} shows the performance of the recognition networks trained with the original data, which shows sufficient performance for evaluating the reconstructed output of the compared encoders by using the recognition models.
For the GSC dataset, we get performance comparable to the 88.2\% in \cite{2018_GSC_dataset}.

\subsection{Intra-Modal Association}

As mentioned in section~\ref{sec:problem_statement}, it is essential for the cross-modal association to learn the intra-modal association that encodes single modal input data into the latent space.
For a fair comparison to existing works, we trained encoders and decoders for each dataset with the fixed dimension of latent space (dim$(\mathbf{z})=64$).
In addition, to show the advantages of the proposed model where the dimension of the latent space can be flexibly designed according to the complexity of target modality, we trained additional auto-encoder whose latent space dimension is $256$ for GTSRB dataset.

Table~\ref{tabular:recog_vae} shows the reconstruction performance of the intra-modal association network implemented by VAE. As shown in the table, the voice data in the GSC dataset shows much degraded accuracy, which means that the voice data are hard to be reconstructed than other modalities.
Since F-MNIST has confused classes such as pullover, coat and shirt, performance on F-MNIST dataset is also degraded.

\input{figure_experiment.tex}

\subsection{Cross-Modal Association}
We evaluated the proposed model on four scenarios: (1) Association between F-MNIST and MNIST, (2) MNIST and GSC, (3) F-MNIST and GSC, (4) GTSRB and GSC.
Scenario (1) is for association between datasets which have similar characteristics to each other.
Scenario (2) and (3) are for association between heterogeneous datasets, i.e. voice and image datasets.
Scenario (4) is a more practical case than the others.
In order to train cross-modal association, we used randomly paired training samples from each dataset belonging to the correlated class.
For example, we paired a randomly chosen sample in '0' class of MNIST dataset with a randomly chosen sample in 'ZERO' class of GSC dataset.

To evaluate the proposed associator, the following methods were compared:
\textbf{VAE} and \textbf{VAE-CG} are variants of the standard VAE~\cite{2013_VAE}. The direct concatenation of paired data is given to VAE as an input. 
To allow VAE to acquire cross-generation capability, \textbf{VAE-CG} is trained to generate the target modality sample via associator for given input sample from other modality.
This model has to be trained only by supervised data with input and output pairs.
Joint Multimodal Variational Auto-Encoder (\textbf{JMVAE})~\cite{2016_joint_multimodal} has two kinds of latent spaces: one is for each modality and the other is for jointly encoding of two modalities. The joint latent space is shared for association between two modalities. The training for encoding in the joint latent space is done to  minimize Kullback-Leibler divergence between the latent vector of each encoder and the joint latent vector of the joint encoder. In comparison, the hyper-parameter $\alpha$ was set to 0.01 for whole scenarios.
Cross-modal Variational Auto-Encoder (\textbf{CVA})~\cite{2018_Handpose} is an extension of VAE for cross-modal data. 
In CVA, the latent space is shared between two modalities. In the training process, the selected sample pair are trained alternately throughout iteration.
Multimodal Variational Auto-Encoder (\textbf{MVAE})~\cite{2018_multimodal_weakly} is also a variant of VAE for cross-modal data. MVAE uses the standard VAE for each modality, but each latent space is associated via a shared latent space expressing the unified distribution of the association modalities. 
We trained MVAE by using the sub-sampled training paradigm presented in their paper.

To evaluate the flexibility of encoding dimension in our model, we have conducted an experiment where each modality is encoded in a different dimensional space from the other.
\textbf{ours-flex} has large dimension of latent space for GTSRB dataset ($dim(\mathbf{z})=256$).
Except for ours-flex, 
all compared models use the same VAE of which the latent space dimension is 64.

Table~\ref{tabular:cross} shows the evaluation result of the proposed model and the compared models for the cross-modal association.
The proposed model accomplishes significant enhancement from the compared algorithms for most of the scenarios.
Interestingly, in the challenging scenarios such as the association between heterogeneous modalities, for instance, between audio (GSC) and visual data (MNIST, GTSRB), the proposed model achieves a remarkable improvement compared to the existing models.

Figure~\ref{fig:experiment} shows the qualitative results of our model for images generated from GSC dataset.
Figure~\ref{fig:experiment} (a) and (b) show 3 generated images for each `number' command of GSC.
Figure~\ref{fig:experiment} (c) shows 5 generated images for each `traffic command' of GSC.
The proposed model successfully generates images with correct semantics though images converge to similar shape due to the random pairing between training samples.

\input{tables/handpose_rebuttal.tex}
\subsection{Application: Hand pose estimation}

We have conducted additional experiments for hand pose estimation on Rendered Hand pose Dataset (RHD)~\cite{2017_RHD_dataset}. 
RHD dataset provides $320 \times 320$ RGB image, depth map, segmentation map and 21 keypoints for each hand.
We consider the case of generating 3D keypoints from the RGB image.
Evaluation metric is the average End-Point-Error (EPE), which measures euclidean distance between ground truth keypoints and estimated keypoints, as CVA~\cite{2018_Handpose} did.
We used the same encoders and decoders structure to CVA and added the associator.
Table~\ref{tabular:handpose} and Figure~\ref{fig:handpose} show that our model outperforms previous works~\cite{2018_Handpose, 2017_RHD_dataset}.
\input{figure_handpose.tex}

\subsection{Semi-supervised Learning}
We conducted an additional experiment to verify the effectiveness of the proposed associator in semi-supervised learning.
Figure~\ref{fig:semi_supervised} illustrates a trend of performance variation of the proposed associator depending on the proportion of paired data, from 100\% to 1\% in the three scenarios including F-MNIST $\rightarrow$ MNIST, MNIST $\rightarrow$ GSC and F-MNIST $\rightarrow$ GSC case.
The result shows that the proposed associator can achieve good performance with only a small proportion of paired data (5\%) in a semi-supervised manner.
\input{figure_semi_supervised.tex}

\subsection{Scalability}

The proposed structure can easily expand a new modality while maintaining the existing modalities. That is, a new modality can be added via training of only a new associator between an existing auto-encoder and the new auto-encoder.
Since the associator connect only two latent spaces, if the existing network associates $N$ modality, $N$ associators need to be trained.
In our model, this inefficiency can be mitigated by cascading association through multiple associators.
For example, suppose that MNIST and F-MNIST datasets are connected by an associator, and GSC and F-MNIST are also connected by an associator. Then even if there is no direct associator between GSC and MNIST, the association between them can be made by the cascading of the two existing associators.
Table~\ref{tabular:scale} compares the results of cascading association and direct association.
Although the cascading association has some performance degradation, it still has good performance compared to other algorithms presented in Table~\ref{tabular:cross}.
\input{tables/scalability_rebuttal.tex}

%% file: tables/cross_results.tex
\begin{table*}[t]
\centering
\caption{Evaluation of cross-modal association models. Accuracy is measured by the recognition model for the reconstructed data of the target modality from the input data of the other modality. Bold font denotes the best performance for each case.
}
\label{tabular:cross}
\begin{tabular}{c|c|c|c|c|c|c|c|c}
\toprule
\multicolumn{9}{c}{Recognition Accuracy (\%)}\\
\midrule
& F-MNIST & MNIST $\rightarrow$ & MNIST & GSC $\rightarrow$ & F-MNIST & GSC $\rightarrow$ & GTSRB & GSC $\rightarrow$ \\
Model&  $\rightarrow$ MNIST & F-MNIST &  $\rightarrow$ GSC	&  MNIST &  $\rightarrow$ GSC &  F-MNIST & $\rightarrow$ GSC & GTSRB \\
\midrule
VAE~\cite{2013_VAE}	        & 47.36 & 41.86 & 10.34 & 28.61 & 12.83 & 22.18 & 35.93 & 19.44\\
VAE-CG~\cite{2013_VAE}  	& 82.41 & 83.84 & 32.46 & 66.62 & 29.87 & 63.79 & 28.43 & 55.00\\
JMVAE~\cite{2016_joint_multimodal}       & \textbf{83.49} & 88.14 & 28.15 & 62.31 & \textbf{47.58} & 51.23 & 41.02 & 65.18\\
CVA~\cite{2018_Handpose}         & 76.51 & 85.88 & 24.61 & 65.04 & 18.70 & 59.73 & 31.02 & 77.78\\
MVAE~\cite{2018_multimodal_weakly}        & 62.65 & 77.62 & 23.04 & 46.70 & 13.52 & 33.24 & 28.06 & 69.17\\
\midrule
\textbf{ours}        & 82.02 & \textbf{94.26} & \textbf{59.47} & \textbf{88.66} & 43.95 & \textbf{77.84} & \textbf{58.89} & \textbf{77.87}\\
\textbf{ours-flex}   & -     & -     & -     & -     & -     & -     & \textbf{61.39} & \textbf{80.56}\\
\bottomrule 
\end{tabular}
\end{table*}

%% file: tables/table_recog_vae.tex
\newcolumntype{L}[1]{>{\raggedright\let\newline\\\arraybackslash\hspace{0pt}}m{#1}}
\newcolumntype{C}[1]{>{\centering\let\newline\\\arraybackslash\hspace{0pt}}m{#1}}
\newcolumntype{R}[1]{>{\raggedleft\let\newline\\\arraybackslash\hspace{0pt}}m{#1}}

\begin{table}[t]
\centering
\caption{Performance of the recognition network trained on the original data and reconstruction performance of VAE. Performance of VAE is measured by the recognition network on the reconstructed data by VAE. dim($\mathbf{z}$) denotes the dimension of latent space of VAE.
}
\vspace{-0.2cm}
\label{tabular:recog_vae}
\begin{tabular}{c|c|cc}
\toprule
Dataset	& Rec (\%) & VAE (\%) & dim($\mathbf{z}$)  \\
\midrule
MNIST	& 97.97 & 96.12 & 64 \\
F-MNIST & 89.22 & 80.54 & 64 \\
GSC		& 88.65 & 81.93 & 64 \\
GTSRB	& 98.53 & 95.70	& 64 \\
GTSRB   & -     & 95.50	& 256 \\
\bottomrule 
\end{tabular}
\vspace{-0.5cm}
\end{table}


%% file: figure_experiment.tex
\begin{figure*}
\centering 
\includegraphics[width=0.85
\linewidth]{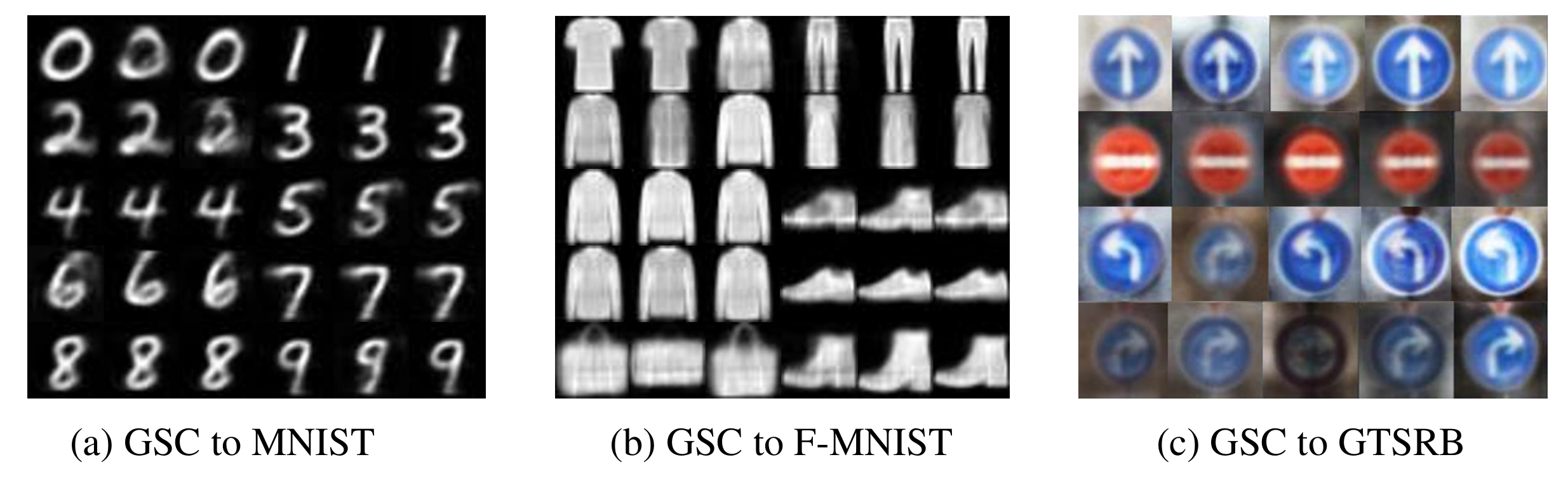}
\caption{Qualitative results on cross-modal association from auditory dataset (GSC) to visual dataset (MNIST, F-MNIST, GTSRB). \textbf{(a)} Image generation from GSC to MNIST dataset. \textbf{(b)} Image generation from GSC to F-MNIST dataset. Results in (a) and (b) shows 3 generated images for each `number' command of GSC.
\textbf{(c)} Image generation from GSC to GTSRB dataset. 
Each row in (c) shows 5 generated images for each `traffic' command of GSC.
}
\label{fig:experiment}
\end{figure*}

%% file: tables/handpose_rebuttal.tex
\begin{table}[t]
\caption{Performance of hand pose estimation on RHD dataset. Performance is measure by average end-point-error.}
\vspace{-0.5cm}
\label{tabular:handpose}
\begin{center}
\begin{tabular}{c|c|c|c}
\toprule
Dataset & \textbf{ours} & CVA~\cite{2018_Handpose} & HPS~\cite{2017_RHD_dataset}\\
\midrule
RHD (RGB $\rightarrow$ 3D) & \textbf{13.15} & 19.73 & 30.42\\                   
\bottomrule 
\end{tabular}
\end{center}
\vspace{-0.7cm}
\normalsize
\end{table}

%% file: figure_handpose.tex
\begin{figure}
\centering 
\includegraphics[width=1.0\linewidth]{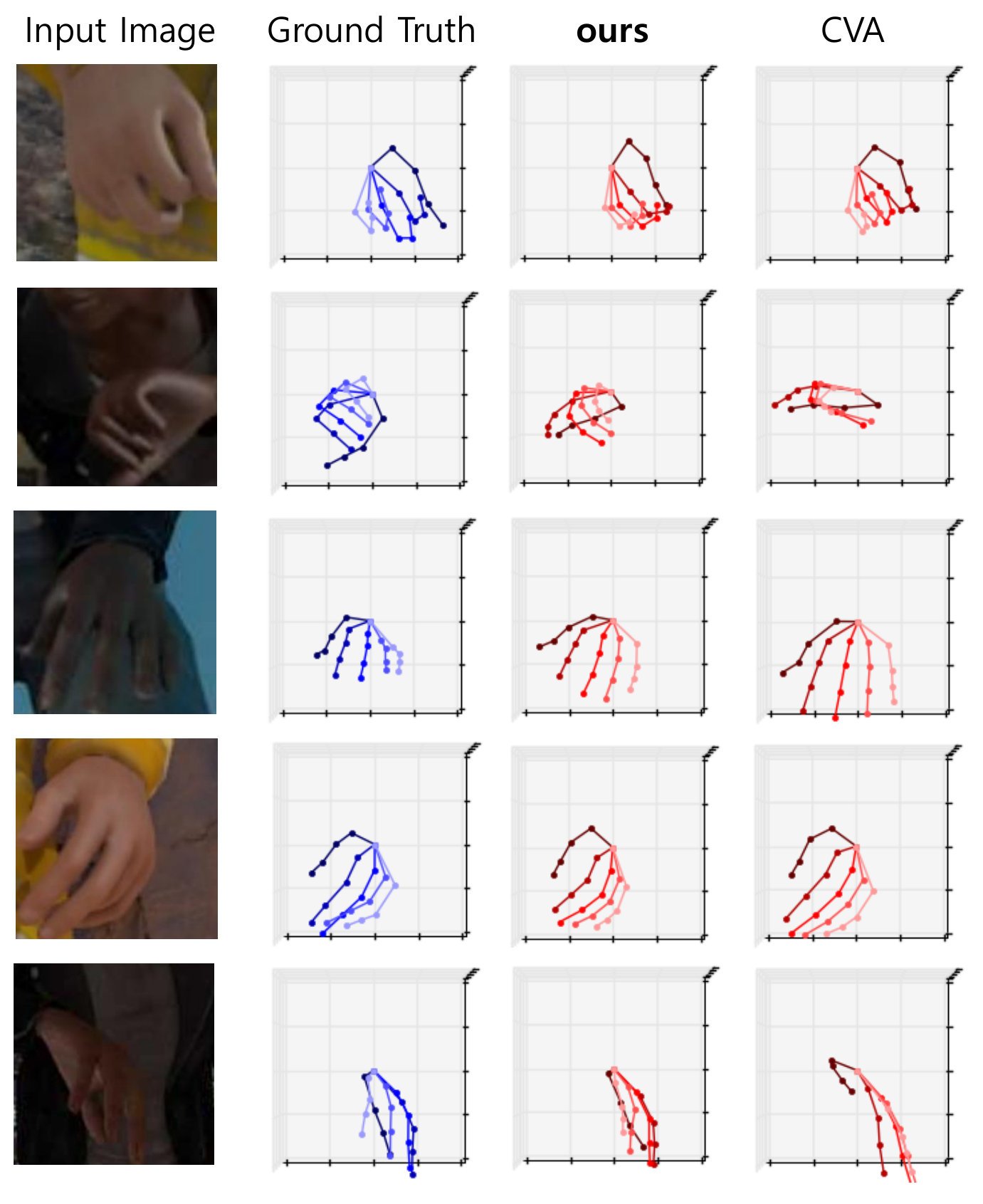}
\caption{Qualitative results for hand pose estimation on RHD dataset.
Each column corresponds to input images, ground truth 3D keypoints, estimated 3D keypoints in order from left to right.
}
\label{fig:handpose}
\end{figure}

%% file: figure_semi_supervised.tex
\begin{figure}
\centering 
\includegraphics[width=0.9\linewidth]{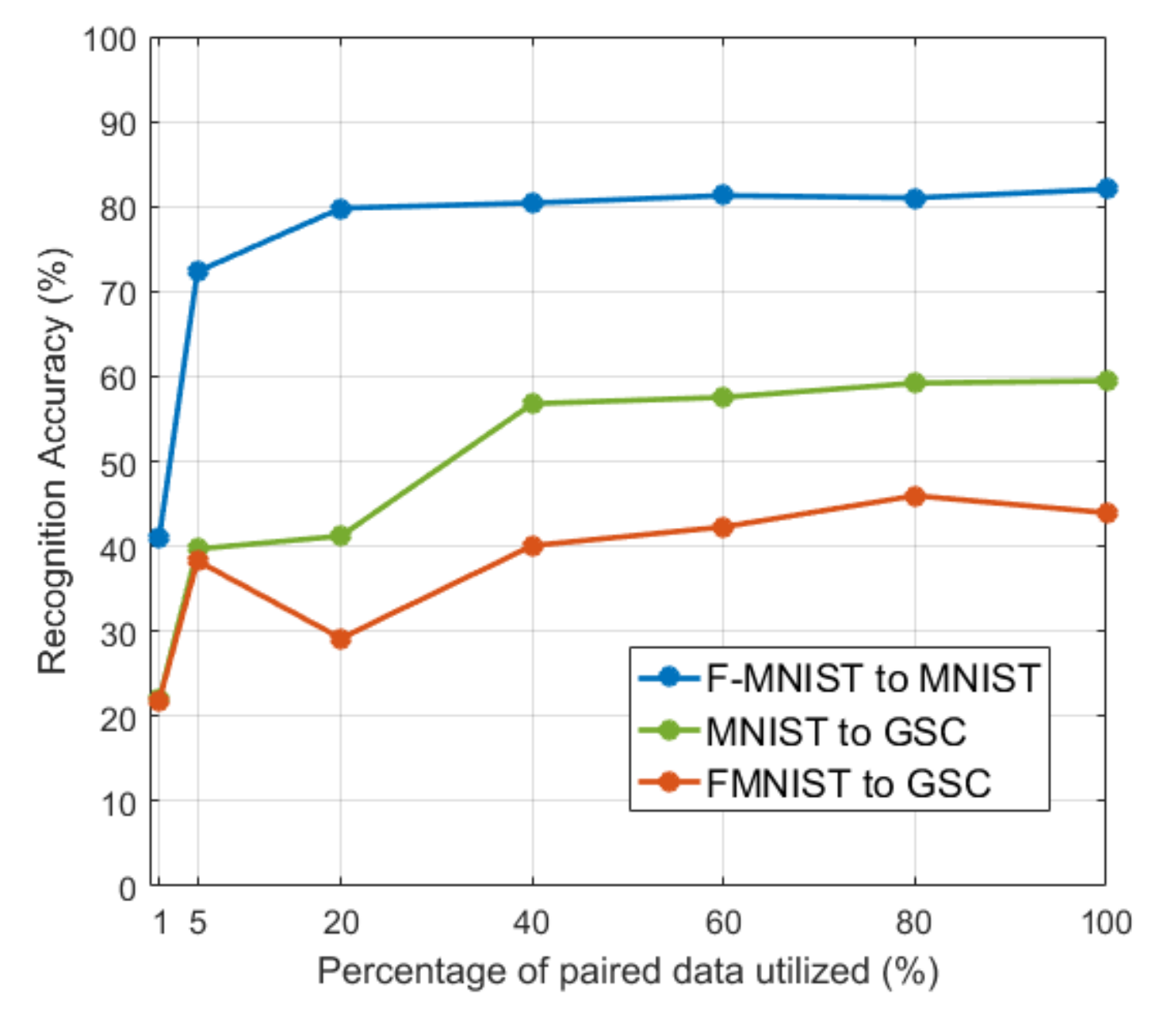}
\caption{\textbf{Semi-supervised learning}. Performance of proposed association network while reducing the proportion of paired data from 100\% to 1\% in the three cases (F-MNIST $\rightarrow$ MNIST, MNIST $\rightarrow$ GSC and F-MNIST $\rightarrow$ GSC).}
\label{fig:semi_supervised}
\end{figure}

%% file: tables/scalability_rebuttal.tex
\begin{table}[t]
\caption{Performance of the proposed model in the case of cascading association and direct association.}
\vspace{-0.5cm}
\label{tabular:scale}
\begin{center}
\begin{tabular}{c|c}
\toprule
GSC $\rightarrow$ MNIST & GSC $\rightarrow$ F-MNIST $\rightarrow$ MNIST\\
\midrule
88.66 & 76.99\\                              
\bottomrule 
\end{tabular}
\end{center}
\vspace{-0.7cm}
\normalsize
\end{table}

%% file: 5-Conclusion.tex
We proposed a novel multi-modal 
association network structure that consists of multiple modal-specific auto-encoders and associators for cross-modal association. 
By adopting the associators, the proposed multi-modal network can incorporate new modalities easily and efficiently while preserving the encoded information in the latent space of each modality.
In addition, the proposed network can effectively associate even heterogeneous modalities by designing each latent space independently and can be trained by a small amount of paired data in a semi-supervised manner. Based on the validation of our structure in experiments, future work can attempt to implement a large-scale multi-modal association network for practical use.

%% file: main.bbl
\begin{thebibliography}{10}\itemsep=-1pt

\bibitem{1985_Boltzmann_Machine}
D.~H. Ackley, G.~E. Hinton, and T.~J. Sejnowski.
\newblock A learning algorithm for boltzmann machines.
\newblock {\em Cognitive science}, 9(1):147--169, 1985.

\bibitem{2018_multimodal_survey}
T.~Baltru{\v{s}}aitis, C.~Ahuja, and L.-P. Morency.
\newblock Multimodal machine learning: A survey and taxonomy.
\newblock {\em IEEE Transactions on Pattern Analysis and Machine Intelligence},
  2018.

\bibitem{2013_representation_survey}
Y.~Bengio, A.~Courville, and P.~Vincent.
\newblock Representation learning: A review and new perspectives.
\newblock {\em IEEE transactions on pattern analysis and machine intelligence},
  35(8):1798--1828, 2013.

\bibitem{1993_synaptic_memory}
T.~V. Bliss and G.~L. Collingridge.
\newblock A synaptic model of memory: long-term potentiation in the
  hippocampus.
\newblock {\em Nature}, 361(6407):31, 1993.

\bibitem{1998_cortical_plasticity}
D.~V. Buonomano and M.~M. Merzenich.
\newblock Cortical plasticity: from synapses to maps.
\newblock {\em Annual review of neuroscience}, 21(1):149--186, 1998.

\bibitem{2016_Robotscene}
C.~Cadena, A.~R. Dick, and I.~D. Reid.
\newblock Multi-modal auto-encoders as joint estimators for robotics scene
  understanding.
\newblock In {\em Robotics: Science and Systems}, 2016.

\bibitem{2017_conditional_multimodal_embedding}
S.~Chaudhury, S.~Dasgupta, A.~Munawar, M.~A.~S. Khan, and R.~Tachibana.
\newblock Conditional generation of multi-modal data using constrained
  embedding space mapping.
\newblock {\em arXiv preprint arXiv:1707.00860}, 2017.

\bibitem{2013_DeViSE}
A.~Frome, G.~S. Corrado, J.~Shlens, S.~Bengio, J.~Dean, T.~Mikolov, et~al.
\newblock Devise: A deep visual-semantic embedding model.
\newblock In {\em Advances in neural information processing systems}, pages
  2121--2129, 2013.

\bibitem{2014_GAN}
I.~Goodfellow, J.~Pouget-Abadie, M.~Mirza, B.~Xu, D.~Warde-Farley, S.~Ozair,
  A.~Courville, and Y.~Bengio.
\newblock Generative adversarial nets.
\newblock In {\em Advances in neural information processing systems}, pages
  2672--2680, 2014.

\bibitem{1961_Hebbian}
D.~O. Hebb.
\newblock {\em The organization of behavior}.
\newblock na, 1961.

\bibitem{2016_Temporal}
D.~Hu, X.~Li, et~al.
\newblock Temporal multimodal learning in audiovisual speech recognition.
\newblock In {\em Proceedings of the IEEE Conference on Computer Vision and
  Pattern Recognition}, pages 3574--3582, 2016.

\bibitem{2013_VAE}
D.~P. Kingma and M.~Welling.
\newblock Auto-encoding variational bayes.
\newblock {\em arXiv preprint arXiv:1312.6114}, 2013.

\bibitem{1998_Lecun_MNIST}
Y.~LeCun, L.~Bottou, Y.~Bengio, and P.~Haffner.
\newblock Gradient-based learning applied to document recognition.
\newblock {\em Proceedings of the IEEE}, 86(11):2278--2324, 1998.

\bibitem{2018_lim_pose}
J.~Lim, Y.~Yoo, B.~Heo, and J.~Y. Choi.
\newblock Pose transforming network: Learning to disentangle human posture in
  variational auto-encoded latent space.
\newblock {\em Pattern Recognition Letters}, 2018.

\bibitem{2000_MFCC_music}
B.~Logan et~al.
\newblock Mel frequency cepstral coefficients for music modeling.
\newblock In {\em ISMIR}, volume 270, pages 1--11, 2000.

\bibitem{2010_MFCC_voicerecognition}
L.~Muda, M.~Begam, and I.~Elamvazuthi.
\newblock Voice recognition algorithms using mel frequency cepstral coefficient
  (mfcc) and dynamic time warping (dtw) techniques.
\newblock {\em arXiv preprint arXiv:1003.4083}, 2010.

\bibitem{2011_MultimodalDeep}
J.~Ngiam, A.~Khosla, M.~Kim, J.~Nam, H.~Lee, and A.~Y. Ng.
\newblock Multimodal deep learning.
\newblock In {\em Proceedings of the 28th international conference on machine
  learning (ICML-11)}, pages 689--696, 2011.

\bibitem{1927_Pavlov}
I.~P. Pavlov.
\newblock Conditional reflexes: an investigation of the physiological activity
  of the cerebral cortex.
\newblock 1927.

\bibitem{1928_Pavlov}
I.~P. Pavlov and W.~Gantt.
\newblock Lectures on conditioned reflexes: Twenty-five years of objective
  study of the higher nervous activity (behaviour) of animals.
\newblock 1928.

\bibitem{2016_MFCC_heartsound}
J.~Rubin, R.~Abreu, A.~Ganguli, S.~Nelaturi, I.~Matei, and K.~Sricharan.
\newblock Classifying heart sound recordings using deep convolutional neural
  networks and mel-frequency cepstral coefficients.
\newblock In {\em Computing in Cardiology Conference (CinC), 2016}, pages
  813--816. IEEE, 2016.

\bibitem{2018_Localize_sound}
A.~Senocak, T.-H. Oh, J.~Kim, M.-H. Yang, and I.~S. Kweon.
\newblock Learning to localize sound source in visual scenes.
\newblock In {\em Proceedings of the IEEE Conference on Computer Vision and
  Pattern Recognition}, pages 4358--4366, 2018.

\bibitem{2018_Handpose}
A.~Spurr, J.~Song, S.~Park, and O.~Hilliges.
\newblock Cross-modal deep variational hand pose estimation.
\newblock In {\em Proceedings of the IEEE Conference on Computer Vision and
  Pattern Recognition}, pages 89--98, 2018.

\bibitem{2012_multimodal_boltzman}
N.~Srivastava and R.~R. Salakhutdinov.
\newblock Multimodal learning with deep boltzmann machines.
\newblock In {\em Advances in neural information processing systems}, pages
  2222--2230, 2012.

\bibitem{2011_German_traffic}
J.~Stallkamp, M.~Schlipsing, J.~Salmen, and C.~Igel.
\newblock The {G}erman {T}raffic {S}ign {R}ecognition {B}enchmark: A
  multi-class classification competition.
\newblock In {\em IEEE International Joint Conference on Neural Networks},
  pages 1453--1460, 2011.

\bibitem{2016_joint_multimodal}
M.~Suzuki, K.~Nakayama, and Y.~Matsuo.
\newblock Joint multimodal learning with deep generative models.
\newblock {\em arXiv preprint arXiv:1611.01891}, 2016.

\bibitem{1999_neurogenesis_mice}
H.~Van~Praag, B.~R. Christie, T.~J. Sejnowski, and F.~H. Gage.
\newblock Running enhances neurogenesis, learning, and long-term potentiation
  in mice.
\newblock {\em Proceedings of the National Academy of Sciences},
  96(23):13427--13431, 1999.

\bibitem{2017_handpose_crossing}
C.~Wan, T.~Probst, L.~Van~Gool, and A.~Yao.
\newblock Crossing nets: Combining gans and vaes with a shared latent space for
  hand pose estimation.
\newblock In {\em Proceedings of the IEEE Conference on Computer Vision and
  Pattern Recognition}, pages 680--689, 2017.

\bibitem{2017_associative_brain}
J.-H. Wang and S.~Cui.
\newblock Associative memory cells: formation, function and perspective.
\newblock {\em F1000Research}, 6, 2017.

\bibitem{2018_associative_brain}
J.-H. Wang and S.~Cui.
\newblock Associative memory cells and their working principle in the brain.
\newblock {\em F1000Research}, 7, 2018.

\bibitem{2016_retrieval_survey}
K.~Wang, Q.~Yin, W.~Wang, S.~Wu, and L.~Wang.
\newblock A comprehensive survey on cross-modal retrieval.
\newblock {\em arXiv preprint arXiv:1607.06215}, 2016.

\bibitem{2018_MixandMatch}
Y.~Wang, J.~van~de Weijer, and L.~Herranz.
\newblock Mix and match networks: encoder-decoder alignment for zero-pair image
  translation.
\newblock In {\em Proceedings of the IEEE Conference on Computer Vision and
  Pattern Recognition}, pages 5467--5476, 2018.

\bibitem{2018_GSC_dataset}
P.~Warden.
\newblock Speech commands: A dataset for limited-vocabulary speech recognition.
\newblock {\em arXiv preprint arXiv:1804.03209}, 2018.

\bibitem{1997_elementary_associative}
E.~A. Wasserman and R.~R. Miller.
\newblock What's elementary about associative learning?
\newblock {\em Annual review of psychology}, 48(1):573--607, 1997.

\bibitem{2004_auditory_cortex}
N.~M. Weinberger.
\newblock Specific long-term memory traces in primary auditory cortex.
\newblock {\em Nature Reviews Neuroscience}, 5(4):279, 2004.

\bibitem{2018_multimodal_weakly}
M.~Wu and N.~Goodman.
\newblock Multimodal generative models for scalable weakly-supervised learning.
\newblock In {\em Advances in Neural Information Processing Systems}, pages
  5580--5590, 2018.

\bibitem{2017_FashionMNIST}
H.~Xiao, K.~Rasul, and R.~Vollgraf.
\newblock Fashion-mnist: a novel image dataset for benchmarking machine
  learning algorithms, 2017.

\bibitem{2017_Yoo_Regression}
Y.~Yoo, S.~Yun, H.~J. Chang, Y.~Demiris, and J.~Y. Choi.
\newblock Variational autoencoded regression: high dimensional regression of
  visual data on complex manifold.
\newblock In {\em Proceedings of the IEEE Conference on Computer Vision and
  Pattern Recognition}, pages 3674--3683, 2017.

\bibitem{2017_RHD_dataset}
C.~Zimmermann and T.~Brox.
\newblock Learning to estimate 3d hand pose from single rgb images.
\newblock Technical report, arXiv:1705.01389, 2017.
\newblock https://arxiv.org/abs/1705.01389.

\end{thebibliography}
